# Forecasting the Olympic medal distribution during a pandemic:

# A socio-economic machine learning model

Christoph Schlembach, Sascha L. Schmidt, Dominik Schreyer, and Linus Wunderlich[*]

**Abstract**

Forecasting the number of Olympic medals for each nation is highly relevant for different stakeholders: Ex ante, sports betting companies can determine the odds while sponsors and media companies can allocate their resources to promising teams. Ex post, sports politicians and managers can benchmark the performance of their teams and evaluate the drivers of success. To significantly increase the Olympic medal forecasting accuracy, we apply machine learning, more specifically a two-staged Random Forest, thus outperforming more traditional naïve forecast for three previous Olympics held between 2008 and 2016 for the first time. Regarding the Tokyo 2020 Games in 2021, our model suggests that the United States will lead the Olympic medal table, winning 120 medals, followed by China (87) and Great Britain (74).

Running head:    Forecasting the Olympic medal distribution

---

[*] Schlembach: WHU – Otto Beisheim School of Management, Erkrather Str. 224a, 40233, Düsseldorf, Germany (e-mail: christoph.schlembach@whu.edu); Schmidt: WHU – Otto Beisheim School of Management, Erkrather Str. 224a, 40233, Düsseldorf, Germany, and CREMA – Center for Research in Economics, Management and the Arts, Switzerland, and LISH – Lab of Innovation Science at Harvard, 175 N. Harvard Street Suite 1350, Boston, MA 02134, USA (e-mail: sascha.schmidt@whu.edu); Schreyer: WHU – Otto Beisheim School of Management, Erkrather Str. 224a, 40233, Düsseldorf, Germany (e-mail: dominik.schreyer@whu.edu); Wunderlich: School of Mathematical Sciences, Queen Mary University London, Mile End Road, London E1 4NS (e-mail: l.wunderlich@qmul.ac.uk).

# 1. Introduction

Forecasting based on socio-economic indicators has a long tradition in academia, in particular in the social sciences. As Johnston (1970, p. 184) noted early, validating the associated "social projections [empirically] may serve to generate appropriate policies or programs whereby we can avoid the pitfalls which would otherwise reduce or eliminate our freedom of action." Consequently, ever since, there have been endeavours to predict the future, exemplary in the field of economics (Modis, 2013), public health (Puertas et al., 2020), civil engineering (Kankal et al., 2011), ecology (Behrang et al., 2011) or urban planning (Beigl et al., 2004).

In the economic literature, in particular, accurately forecasting Olympic performances has gained considerable research interest over the last decades (cf. Leeds, 2019), primarily because such forecasts, typically medal forecasts, are necessary to provide both a government and its citizens with a benchmark against which they can evaluate the nation's Olympic success ex-post. For a government, often investing heavily in athlete training programs to enhance the probability of a nation's Olympic success (cf. Humphreys et al., 2018), such an assessment is pivotal because it allows them to understand better whether the application of funds, i.e. the taxpayers' money, to their National Olympic Committee (NOC) is productive. For instance, as major sporting events such as the Olympic Games are often associated with increasing both national pride among their citizens (cf. Ball, 1972; Grimes et al., 1974; Allison and Monnington, 2002; Hoffmann et al., 2002; Tcha and Pershin, 2003) and their willingness to begin engaging in sporting activities (cf. Girginov and Hills, 2013; Weed et al., 2015), thereby reducing long-term healthcare costs, a government might be willing to raise funds if their NOC meets (or exceeds) the medal forecasts. In contrast, because Olympic success is a well-known antecedent of civic willingness to support funding a government's elite athlete training programs (Humphreys et al., 2018), falling behind the predictions might motivate a government to increase the pressure on the NOC, not least by reducing future funds.

Likewise, accurately forecasting the Olympic success is highly relevant for many non-governmental different stakeholders. For instance, sports betting companies rely on precise estimates to determine their odds, while both the media and Olympic sponsors must allocate their resources to promising teams and their athletes. Thus, analysing the Olympic Games empirically has become a relevant field of research, both, with a focus on forecasting (e.g. De Bosscher et al., 2006) and beyond (e.g. Streicher et al., 2020).

Since the first contribution by Ball (1972), the quality of such Olympic forecasts has steadily improved for two reasons. First, those authors interested in predicting a nation's Olympic success have successively begun employing new estimation techniques. Second, over time, the predictive power of models has gradually increased as authors operating in the field have explored diverse, increasingly extensive data sets.

Since Ball (1972) pioneered with a correlation-based scoring model, forecasting models have continuously become more sophisticated. Initially, as we show in Table A1 in the appendix, most authors referred to the use of ordinary least squares regressions (OLS), as it delivered results that were easy to interpret (e.g. Baimbridge, 1998; Condon et al., 1999; Kuper and Sterken, 2001). However, a significant challenge when predicting Olympic medals is to reflect the large number of nations without any medal success properly. As the incorporated exponential function punishes small predicted numbers of medals, some authors (e.g. Lui and Suen, 2008; Leeds and Leeds, 2012; Blais-Morisset et al., 2017), then, moved to Poisson-based models (i.e., a Poisson model, negative binomial model), to tackle this methodological problem. However, until today, because the dependent variable, typically the number of medals, is censored, most authors have employed Tobit regression to predict Olympic success (e.g. Tcha and Pershin, 2003; Forrest et al., 2015; Rewilak, 2021). Only recently, employing a two-step approach, estimating the probability of winning any medal before determining the exact number of medals in case of success, became more popular. In particular, both Scelles et al. (2020) and Rewilak (2021), employing a Mundlak transformation of the Tobit model, could, again, increase the prediction accuracy with their respective Hurdle models. In contrast, other authors, e.g. Hoffmann et al. (2002), circumvent the underlying methodological problems by splitting their sample into nations who did and did not win any medals in the past, while few authors have employed alternative methodological approaches.[1] However, despite these methodological improvements, a naïve forecast still outperforms all these previous forecasting approaches regularly.

Somewhat similarly, during the last years, authors have significantly increased the data sets used for medal forecasting in three ways. First, by increasing the level of granularity beyond country-specifics; second, by including more years; and third, by exploring additional independent variables.

As a common way to incorporate more granular data, and thus to increase the forecast accuracy, some authors considered predicting the Olympic success by focussing on different sports (e.g. Tcha and Pershin, 2003; Noland and Stahler, 2016a; Vagenas and Palaiothodorou, 2019), sometimes even exploring data on the level of the individual athlete (Condon et al., 1999; Johnson and Ali, 2004). Due to the increasing relevance of gender studies, other authors have begun differentiating their data sets by gender (Leeds and Leeds, 2012; Lowen et al., 2016; Noland and Stahler, 2016b). Noticeably, both approaches are certainly important to answer very specific questions. Yet, macro-level models, in contrast, have the "advantage of averaging the random component inherent in individual competition [leading to] more accurate predictions of national medal totals" (Bernard and Busse, 2004, p. 413). Thus, macro-level analysis remains a frequently used approach in Olympic medal forecasting.

A different approach to potentially increase the forecast accuracy is to expand the data set's temporal dimension. As such, some authors have incorporated one hundred years of Olympics and more in their models (e.g. Baimbridge, 1998; Kuper and Sterken, 2001; Trivedi and Zimmer, 2014). However, more recently, most authors seem to limit the number of events under investigation for three reasons. First, specific incidents such as "large-scale boycotts as occurred at the 1980 Moscow and 1984 Los Angeles Games" (Noland and Stahler, 2016b, p. 178) and "the East German doping program [, which] was responsible for 17 percent of the medals awarded to female athletes" (Noland and Stahler, 2016b, p. 178) in 1972 skewed the medal count in the past. Second, frontiers shifted particularly in the course of two World Wars and the breakdown of the Soviet Union; only since the 1990s nations remained relatively stable (Forrest et al., 2017). Third, the significance of variables changed over time, such that, for instance, a potential host effect, might play a different role in times where international travel has become a part of our daily lives (Forrest et al., 2017).

---

[1] For instance, Condon et al. (1999) employed neural networks for one Olympiad. Other authors used (single-step) binary regression models (Probit, Logit) while using more granular data (Johnson and Ali, 2004; Andreff et al., 2008; Noland and Stahler, 2016b).

Finally, another way to augment a data set and, thus, to improve the accuracy of a forecast is to incorporate additional independent variables. Early, Ball (1972) found that "[g]ame success is related to the possession of resources, both human and economic, and the centralized forms of political decision-making and authority which maximize their allocation" (p. 198). Particularly, the extraordinary Olympic performance of countries with a certain political system, at that time the Soviet Union, has been confirmed by research until today, e.g. Scelles et al. (2020). Other authors (cf. Kuper and Sterken, 2001; Hoffmann et al., 2002; Bernard and Busse, 2004; Johnson and Ali, 2004) found that hosting the Games increases the expected number of medals, among others due to an increased number of fans and reduced stress due to international travel. Maennig and Wellbrock (2008), Forrest et al. (2010) and Vagenas and Vlachokyriakou (2012) extended this finding and concluded that such a host effect already starts four years before the Olympic Games and, surprisingly, lasts until the subsequent Games. On a similar note, authors found a continuous over-, respectively underperformance of nations, such that lagged medal shares significantly improve the prediction accuracy (cf. Bernard and Busse, 2004).[2]

In addition, it is important to mention that many scholars experimented with additional variables such as climate (Hoffmann et al., 2002; Johnson and Ali, 2004), public spending on recreation (Forrest et al., 2010), health expenditure, growth rate, unemployment (Vagenas and Vlachokyriakou, 2012), and income (Kuper and Sterken, 2001). In general, there are mixed findings on most of these variables and only few are available in public as comprehensive data sets. In this regard, De Bosscher et al. (2006) conducted a meta-analysis of variables predicting sportive success, even beyond Olympic Summer Games, and found that both the Gross National Product and the country's population "consistently explain over 50% of the total variance of international sporting success" (p. 188). It is, therefore, not surprising that these two variables, in particular, have been used by most authors in the past 50 years. As such, also taking the potential issue of multicollinearity from exploring too many distinct, though potentially related, socio-economic variables into consideration, forecasting models should not be augmented infinitely.

Given the high policy relevance and academic attention of Olympic forecasting, it is somewhat surprising that the potential of machine learning in detecting hidden patterns and, thus, improving forecasting accuracy has not yet been exhausted in this context. However, this methodology has recently received an increasing level of popularity in a sports context, e.g. in football (Baboota and Kaur, 2019). Particularly the Random Forest approach often delivers excellent results, for instance, in forecasting football scores (Groll et al., 2019) or horseracing outcomes (Lessmann et al., 2010). As acknowledged by Makridakis et al. (2020), statistical knowledge can be applied in the world of machine learning as well.

---

[2] This also explain why none of the published models repeatedly outperformed a naïve forecasting model assuming the number of medals from the preceding Games for the upcoming Games as well.

As such, in this study, we translate the proven concept of the Tobit model to machine learning by using a two-staged Random Forest model to predict Olympic performance. In that way, we identify the first model to consistently outperform a naïve forecasting model, in three consecutive Summer Games (2008, 2012, and 2016) by about 3 to 4 percentage points. On a side note, we thus also improve the forecasting accuracy presented in more recent work on the potential determinants of Olympic success (Scelles et al., 2020) by roughly 20 percent.[3]

The remainder of our manuscript is structured as follows: After motivating the variables used in the model and introducing the concept of a two-staged Random Forest, we evaluate the quality of the forecast, present an estimate for Tokyo 2020, and discuss the implications of COVID-19. We conclude with a summary, ex-ante and ex-post consequences of the prediction, and an outlook for further research.

## 2. Material and Methods

We forecast the number of medals in the Tokyo Olympic Games for each participating nation based on a two-staged Random Forest. It is, however, important to note that, as part of this exercise, we also quantify the impact of COVID-19 on the expected Olympic medal count based on the independent variables (i.e., features) national GDP, incidents of and deaths from lower respiratory diseases. In this section, we motivate the underlying variables, explain the concept of a two-staged Random Forest, and describe the forecasting process.

*2.1 Variables*

*2.1.1 Dependent variable (output variable)*

---

[3] The significant increase in predictive accuracy is based on two effects: First, Scelles et al. (2020) build on a generalized linear model in form of a Hurdle, respectively Tobit, model. We, in contrast, apply a two-staged Random Forest algorithm taking into account more complex, non-linear interactions. Second, it is often argued that the time to prepare an Olympic team is four years (cf. Forrest et al. (2010); Scelles et al. (2020). This would imply that, ideally, only socio-economic data until 2016 should be used to predict the Tokyo 2020 results. However, Stekler et al. (2010) evaluate different sports forecasting methodologies and find that more recent data generate better results. Thus, we include data until 2020 in our model to overcome this issue, which is even amplified by the WHO's decision to declare COVID-19 a pandemic (Cascella et al. (2020) and the subsequent postponement of the Games to 2021 (International Olympic Committee (2020).

The number of Olympic medals represents economic and political strength and promotes national prestige (Allison and Monnington, 2002). De Bosscher et al. (2008, p. 19) acknowledge the figure as "the most self-evident and transparent measure of success in high performance sport". As most scholars (e.g. Andreff et al., 2008; Scelles et al., 2020), we define the number of medals as dependent variable without distinguishing between gold, silver and bronze medals. Following Choi et al. (2019), who note that a log-transformation can reduce the skewness and, thus, improve prediction accuracy in machine learning, we take the logarithm of the number of medals, which reduces the (right-) skewness from 3.2 to 0.4 (i.e., only among non-zero medals due to the definition of logarithm). As the independent variables do not change at the same rate as the Olympic medal totals, we cannot expect an exact match between forecast and actual medals at stake. Thus, we need to rescale the prediction to the number of scheduled events times three (assuming no double bronze). Further, rounding is necessary to get natural numbers.

*2.1.2 Independent variables (features)*

The predictive power of GDP for sportive success is widely accepted in Olympic medal forecasting (cf. Bernard and Busse, 2004) and robust across both geographies (cf. Manuel Luiz and Fadal, 2011) and sports (e.g. Klobučník et al., 2019)[4]. Several reasonable explanations include that richer nations invest higher sums in sports, provide more extensive sports offerings and cater for a better overall fitness among the population (De Bosscher et al., 2008). Due to limited data availability and high allocation complexity on a more granular level, aggregate figures, such as GDP, have become a de facto standard in academia (De Bosscher et al., 2006; Manuel Luiz and Fadal, 2011). To account for the character of the Olympic Games as a competition, we normalize and use the share of a nation in the global GDP as feature.

---

[4] Even though the extent research on predicting Olympic success typically suggests that the total GDP "is the best predictor of national Olympic performance" (Scelles et al. , 2020, p. 698), we have further experimented with a number of both alternative and additional independent variables capturing a nation's economic status. For instance, to proxy the degree of inequality and poverty in a certain country, we have also added information based on the Human Development Index (HDI) to our model, that is, a statistic composite index of life expectancy, education, and per capita income indicators, which are frequently employed to rank countries into four tiers of human development. However, we did not find any improvements in prediction accuracy. These additional results are available from the first author upon request.

Besides GDP, the population of a nation is a well-established predictor of Olympic medals (Bernard and Busse, 2004; De Bosscher et al., 2008); that is, larger countries have larger resources of potential medal winners (Bernard and Busse, 2004). As the number of world-class athletes in a country, however, is exhausted at some point and population alone does not lead to more medals anymore[5], we take the logarithm of the population which grows slower than a linear function.

**Table 1.**

Descriptive statistics of numerical variables used in the model including data sources.

| Variable | Type | Minimum | Maximum | Mean | Std. Deviation | Skewness | Data Source |
|---|---|---|---|---|---|---|---|
| Number of medals | Numerical | 0 | 121 | 4.639 | 13.190 | 5.024 | Griffin (2018) |
| Share of global GDP | Numerical | <0.001 | 0.200 | 0.005 | 0.017 | 7.773 | International Monetary Fund, 2019, 2020; The World Bank, 2020 |
| Population (E+8) | Numerical | 2.287 | 14.157 | 8.398 | 2.275 | -0.512 | United Nations, Department of Economic and Social Affairs (2019) |

*Abbreviations and notes.* We display all values from 1991 to 2016 as 2020 medals are not known yet.

We also reflect the number of participating athletes in the model. Scelles et al. (2020) suggest the use of categorical variables for the number of athletes. Here, the rationale is that the final number of competitors is generally not known at the time of forecasting. Furthermore, the categories suggested by them have rarely changed in the past. As an example, Afghanistan has always sent between zero and nine athletes since 1992. Before forming these groups, we count athletes that started in multiple disciplines multiple times as their chances to win a medal multiply.

While existing research confirms an impact of specific socio-economic variables on the number of medals in the Olympic Games, the connection of public health crises and sportive performance was not to be presumed before the COVID-19 crisis. Yet, this pandemic did not only lead to the postponement of the Games to 2021 (International Olympic Committee, 2020) but also affected the athletes' preparation (Mohr et al., 2020; Mon-López et al., 2020; Wong et al., 2020), as well as the funds available in the sports industry (Hammerschmidt et al., 2021; Horky, 2021; Parnell et al., 2021). We reflect the impact of COVID-19 via incidents of and deaths from lower respiratory diseases, as well as GDP. We categorise the two first-mentioned features in quintiles to limit the effect of potential outliers. The broad availability of data allows us to create a synthetic "no COVID-19" scenario by eliminating COVID-19 incidents and deaths, and by leveraging GDP forecasts made before the beginning of the pandemic; hence, we can quantify the impact of COVID-19 on Olympic medals.

---

[5] For instance, in India, the world's second largest country based on population, the significant growth of population between 1992 and 2016 (CAGR: 1,58%) was hardly converted into Olympic medals; while India won zero medals in 1992, the number did not increase significantly until 2016 (two medals). As a comparison, China, a country with a similar population, won 70 medals in 2016. For an in-depth analysis of the Olympic performance of India, please refer to Krishna and Haglund (2008).

Already Ball (1972, p. 191), in his seminal contribution, mentioned that "hosts [of Olympic Games] are more successful, at least in part because of their ability to enter larger than usual teams at relatively low financial expenditure". Social pressure on the decision making of participants might also explain such a "host-effect" (Garicano et al., 2005; Dohmen and Sauermann, 2016; Bryson et al., 2021). We, therefore, include a categorical variable for past, current, and future host countries.

Bernard and Busse (2004) detected, that Soviet countries outperformed their expected medal success on a regular basis due to the essential role of sports in the communist regimes. Starting early on and combining competitive sports and education was an essential component in their strategy (Metsä-Tokila, 2002). Reflecting such peculiarities in political systems, we use the trichotomy in capitalist market economies, (post-) communist economies and Central Eastern European countries, that joined the EU, as refined by Scelles et al. (2020).

Further, geographic characteristics determine the capabilities of succeeding in a given sport because of culture, tradition and climate (Hoffmann et al., 2002). Subsequently, we use 21 regions as defined by the United Nations, Department of Economic and Social Affairs (2020) as categorical independent variable.

Finally, as recommended by Scelles et al. (2020) and Celik and Gius (2014), the number of medals in the preceding Olympics (non-logarithmic) is added to the model as it significantly improves the predictive power. This suggests that there are some unconsidered country-specific factors, which may "include a nation's athletic tradition, the health of the populace, and geographic or weather conditions that allow for greater participation in certain athletic events" (Celik and Gius, 2014, p. 40).

We display the descriptive statistics of the numerical variables used in the model in Table 1 and list ordinal and categorical variables in Table 2. As a rule of thumb, Stekler et al. (2010) find that more recent data generate better results in sports forecasting. Thus, we leverage data from one year prior to the Olympics to make a forecast. Thus, being able to retrieve data points of 206 countries between 1991 and 2020, we can feed our models with 1,379 country-year observations.

**Table 2.**

List of ordinal and categorical variables used in the model including data sources.

| Variable | Type | Number (ones) | Data Source |
|---|---|---|---|
| Number of athletes | Ordinal | | Griffin, 2018; Scelles et al., 2020 |
|    0-9 Athletes | | 589 | |
|    10-49 Athletes | | 388 | |
|    50-149 Athletes | | 230 | |
|    Over 149 Athletes | | 172 | |
| Diseases Deaths (deaths due to lower respiratory diseases) | Ordinal (quintiles) | | Global Burden of Disease Collaborative Network (2018) |
| Diseases Incidents (people affected by lower respiratory diseases) | Ordinal (quintiles) | | Global Burden of Disease Collaborative Network (2018) |
| Deaths due to COVID-19 | Ordinal (added to Diseases Deaths) | | Institute for Health Metrics and Evaluation, 2020; World Health Organization, 2020 |
| COVID-19 incidents | Ordinal (added to Diseases Incidents) | | Institute for Health Metrics and Evaluation, 2020; World Health Organization, 2020 |
| Host country | Categorical | | Wikipedia (2020) based on the International Olympic Committee |
|    Current Host | | 7 | |
|    Last Time's Host | | 7 | |
|    Next Host | | 7 | |
| Political regime | Categorical | | Scelles et al. (2020) |
|    CAPME (capitalist market economies) | | 1,161 | |
|    POSTCOM ((post-) communist economies) | | 141 | |
|    CEEC, joined the EU (Central Eastern European countries) | | 77 | |

| Variable | Type | Number (ones) | Data Source |
|---|---|---|---|
| Region | Categorical | | United Nations, Department of Economic and Social Affairs (2020) |
|     Sub-Saharan Africa | | 314 | |
|     Latin America & Caribbean | | 263 | |
|     Western Asia | | 122 | |
|     Southern Europe | | 95 | |
|     South-eastern Asia | | 72 | |
|     Northern Europe | | 70 | |
|     Eastern Europe | | 67 | |
|     Western Europe | | 63 | |
|     Southern Asia | | 61 | |
|     Eastern Asia | | 49 | |
|     Northern Africa | | 42 | |
|     Polynesia | | 31 | |
|     Central Asia | | 30 | |
|     Micronesia | | 30 | |
|     Melanesia | | 28 | |
|     Northern America | | 21 | |
|     Australia and New Zealand | | 14 | |
|     Western Africa | | 7 | |

*2.2 Data pre-processing*

Data pre-processing is a vital step to ensure accurate forecasts (Wang et al., 2018; Chen et al., 2019). Here, we perform three steps: First, mapping of nations; second, inter-/extrapolation; and third, benchmarking.

*2.2.1 Mapping of nations*

As Olympic teams according to the definition of the International Olympic Committee do not necessarily match the country list in other data sources, we need to (dis-) aggregate socio-economic data to adequately represent Olympic teams, e.g. by adding the population of Anguilla, a part of Great Britain, to the British population as accounted in the data source. In 2016, we attribute nine International Olympic Athletes (IOA) winning two medals to Kuwait based on their nationality. The Unified Team (EUN) represents Russia being banned from the Games because of doping (Hermann, 2019). A "neutral" team might participate in 2021 again; in this case our prediction applies to athletes from Russia regardless the name of their team. We split the athletes (269) and medals (7) of the former Czechoslovakia into Czech Republic (178

/ 5) and Slovakia (91 / 2) based on their respective population; this allows adequate forecasts for the two nations that emerged from Czechoslovakia. The Refugee Olympic Team (12 athletes in 2016) has not won a medal yet. Hence, we assume a constant forecast meaning that there will be no medals in 2021 either.

*2.2.2 Inter- / extrapolation*

We obtain missing data points in a specific year by inter- / extrapolation, which is a common approach when pre-processing data (e.g. Christodoulos et al., 2010; Chen et al., 2019). While we interpolate linearly if certain years between two data points are missing, we extrapolate by assuming constant numbers based on the first (respectively last) value in the dataset; this way we do not mis-interpret local events (Scott Armstrong and Collopy, 1993). Exceptions are GDP and population which typically inhibit a consistent growth; linear extrapolation is sensible here. For nations, where not more than five consecutive points are missing and there are more data points available than missing, we extrapolate linearly to take the implicit trend into account by using a constrained least-squares approach: With $n<6$ missing values, we use the $n+1$ nearest available values to estimate the slope of the line. The intercept is given by the nearest available value.

*2.2.3 Benchmarking*

If there are no data points for one country available at all, we leverage the average of the respective region (United Nations, Department of Economic and Social Affairs, 2020) as benchmark. The rationale here is that countries within one region also share socio-economic characteristics, such as economic strength. Yet, this approach is only necessary for some variables of some smaller countries, which are responsible for 1% of total Olympic medals.

*2.3 Conceptual development*

The Tobit model marked a milestone in Olympic medal forecasting accounting for the large number of nations winning zero medals (Bernard and Busse, 2004). The concept traces back to Tobin (1958), who argues that for censored variables, linear regression models do not deliver suitable results. To apply this statistical concept in machine learning, we develop a two-staged algorithm: As a first step, we train a binary classifier to determine whether a nation should win any medals or not. As a second step, we train a regression model to forecast the exact number of medals for countries with predicted medal success.

In both steps, we employ a Random Forest algorithm (cf. Lee, 2021), an *ensemble learner* which has been proven advantageous in various disciplines of sports forecasting (cf. Lessmann et al., 2010; Groll et al., 2019). Breiman (2001, p. 5) explains that "Random [F]orests are a combination of tree predictors such that each [decision] tree depends on the values of a random vector sampled independently and with the same distribution for all trees in the forest" and reports "significant improvements in classification accuracy". The difference between classifiers, as used in the first step of our model, and regressions, as used in the second step, is that the tree predictor of classifiers reports class labels ("zero medals" or "medal success") as opposed to numerical values describing the number of medals (Breiman, 2001; Cutler et al., 2012; Lee, 2021). While classifiers use majority votes of the individual decision trees, regression models average the values determined by the trees. When setting up the number of trees in an ensemble, Oshiro et al. (2012) find that larger trees do not necessarily improve the performance. Thus, we use ten trees in the first step. To provide meaningful confidence intervals for the final estimates (based on ensembles of ten trees), we use one thousand trees in the second step.

Cutler et al. (2012) explains why Random Forests are appealing from both, a computational (among others due to training and prediction time, small number of parameters, and direct use for high dimensional problems) and a statistical (among others due to measures of variable importance, differential class weighting and outlier detection) point of view. The main shortcoming of (individual) decision trees is that they are prone to overfitting (Kirasich et al., 2018). Even though Random Forests partly account for this issue "by using a combination or 'ensemble' of decision trees where the values in the tree are a random, independent, sample" (Kirasich et al., 2018, p. 7), a diligent setup of the forecasting process is essential.

*2.4 Forecasting process*

For this reason, we apply a rigorous time-consistent data separation to avoid cases of overfitting and obtain impartial and robust results (Dwork et al., 2015; Roelofs et al., 2019). *Cross-validation* splits the dataset in training and testing data; while the training data determines the model, the testing data ensures validity beyond a fixed data sample (Kerbaa et al., 2019; Li et al., 2020). In case of time series, as presented in this paper, it is essential to consider the temporal evolution and dependencies in the data; therefore, Bergmeir and Benítez (2012) suggest *last block cross-validation*, a special case of cross-validation using the most recent datapoints as testing data.

More specifically, we use data collected from the years 1991 to 2004 as the *training set*, and data from the 2008 Olympic Games as the *test set*, to evaluate and compare the performance of distinct models. Only then, we evaluate the final model on the *validation set*, which includes data of the 2012 and 2016 Olympic Games. We illustrate the overall forecasting process in Figure 1.

To demonstrate the performance of the previously introduced two-staged Random Forest, we benchmark the model against others based on the share of nations with correctly forecast medals in the 2008 Olympic Games (*training of different models*). For the first step of the model, as classifier, we also consider a support vector machine and Random Forests with one, a hundred, and a thousand trees. For the second step, the regression, we benchmark against a range of classical regressions, boosting methods, and neural networks: As classical regressions, we consider a linear regression, a Support Vector Machine taking into account non-linear transformations of features (Chang and Lin, 2011), and a decision tree regression (Breiman et al., 1984). Boosting methods (Bühlmann and Hothorn, 2007) perform several instances of decision trees (in this case). Each tree compares the output variable to the forecast from the previous step and adapts the setup for the next step based on the error. We include AdaBoost (Freund and Schapire, 1997), which directly takes into account the error, and XGBoost (Chen and Guestrin, 2016), which first transforms the error, as benchmark. Neural networks (LeCun et al., 2015) were motivated by the structure of a biological brain. They use a computational network, where each node performs a simple transformation and hands over the result to subsequent nodes. In both steps, as classifier (10 trees) and regression model (1,000 trees), the Random Forest outperforms the described algorithms.

To further validate this performance, we use the two-staged Random Forest to forecast the 2012 and 2016 Olympic Games (*model validation*). This is particularly interesting as it allows us to compare our model to past medal forecasts presented in academic literature. When computing our estimates, we use the same datapoints that had already been available when the respective papers were developed.

Finally, we make predictions for the 2020 Olympic Games (*forecasting*); this time, we use data containing the 2016 Olympics as well. While we show these results in Table 4, none of the additional information, i.e. data on the Olympics 2016, is used in the forecasts summarized in Table 3.

We implement all models in Python 3.8.5 (Oliphant, 2007) using the packages pandas 1.1.2 (McKinney, 2010), scikit-learn 0.23.2 (Pedregosa et al., 2011), XGboost 1.2.0 (Chen and Guestrin, 2016), NumPy 1.18.5 (van der Walt et al., 2011), and Shap 0.36.0 (Lundberg and Lee, 2017).

**3. Results**

While Scelles et al. (2020) improved the prior forecast quality, the presented models still fail to outperform a naïve forecast, i.e. assuming that each country wins exactly the same number of medals as in the previous Olympics. The approach presented in this paper is, to the best of our knowledge, the first to consistently beat the naïve forecast for the 2008, 2012, and 2016 Games (cf. Table 3). Besides the naïve forecast, we include five other models from three different papers in our benchmark. We evaluate the forecast accuracy, i.e. the share of correctly predicted medals, in total and for all nations winning zero, respectively more than zero medals. Furthermore, we check whether the forecast lies in a 95% confidence interval augmented by two medals. Finally, we sum the absolute deviation of forecast and actual medals of the top-17 nations as suggested by Scelles et al. (2020). Note that our sample includes 203 (in 2008), 205 (in 2012), respectively 206 (in 2016) participating nations, while Forrest et al. (2010) model 127, Andreff et al. (2008) model 20, and Scelles et al. (2020) model 192 nations. When computing our estimates, we use the same datapoints that had already been available when the respective papers were developed (cf. Figure 1).

**Figure 1.**

Illustration of forecasting process.

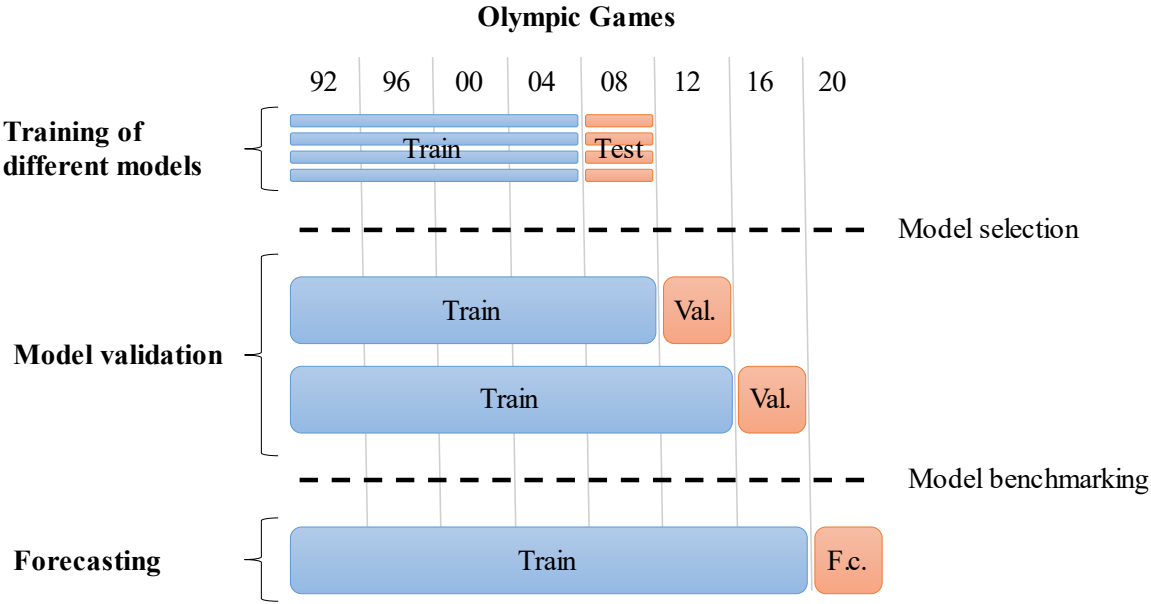

*Abbreviations and notes.* Val. = Validation Set, F.c. = Forecast

**Table 3.**

Forecasting accuracy of selected models.

|  | 2008 | 2012 | 2016 |
|---|---|---|---|
| Correct forecast | | | |
|     Two-Staged Random Forest (This Paper) | 63% | 59% | 64% |
|     Naïve Forecast | 59% | 56% | 60% |
|     Tobit Model (Forrest et al., 2010) | 47% | | |
|     Tobit Model (Andreff et al., 2008) | 5% | | |
|     Logit Model (Andreff et al., 2008) | 0% | | |
|     Hurdle Model (Scelles et al., 2020) | | | 22% |
|     Tobit Model (Scelles et al., 2020) | | | 43% |
|     Tobit Model (Maennig and Wellbrock, 2008) | 41% | | |
|     OLS (Celik and Gius, 2014) | | 10% | |
| Correct forecast (non-zero medals) | | | |
|     Two-Staged Random Forest (This Paper) | 14% | 11% | 17% |
|     Naïve Forecast | 9% | 11% | 16% |
|     Tobit Model (Forrest et al., 2010) | 17% | | |
|     Tobit Model (Andreff et al., 2008) | | | |
|     Logit Model (Andreff et al., 2008) | | | |
|     Hurdle Model (Scelles et al., 2020) | | | 22% |
|     Tobit Model (Scelles et al., 2020) | | | 11% |
|     Tobit Model (Maennig and Wellbrock, 2008) | 11% | | |
|     OLS (Celik and Gius, 2014) | | 10% | |
| Correct forecast (zero medals) | | | |
|     Two-Staged Random Forest (This Paper) | 98% | 93% | 97% |
|     Naïve Forecast | 96% | 88% | 92% |
|     Tobit Model (Forrest et al., 2010) | 94% | | |
|     Tobit Model (Andreff et al., 2008) | | | |
|     Logit Model (Andreff et al., 2008) | | | |
|     Hurdle Model (Scelles et al., 2020) | | | 22% |
|     Tobit Model (Scelles et al., 2020) | | | 69% |
|     Tobit Model (Maennig and Wellbrock, 2008) | 83% | | |
|     OLS (Celik and Gius, 2014) | | | |
| 95% confidence intervals +/- 2 medals | | | |
|     Two-Staged Random Forest (This Paper) | 92% | 96% | 93% |
|     Naïve Forecast | | | |
|     Tobit Model (Forrest et al., 2010) | | | |
|     Tobit Model (Andreff et al., 2008) | 60% | | |
|     Logit Model (Andreff et al., 2008) | 45% | | |
|     Hurdle Model (Scelles et al., 2020) | | | 93% |
|     Tobit Model (Scelles et al., 2020) | | | 91% |
|     Tobit Model (Maennig and Wellbrock, 2008) | | | |
|     OLS (Celik and Gius, 2014) | | | |
| Absolute deviation top-17 nations | | | |
|     Two-Staged Random Forest (This Paper) | 152 | 91 | 128 |
|     Naïve Forecast | 154 | 115 | 114 |
|     Tobit Model (Forrest et al., 2010) | 92 | | |
|     Tobit Model (Andreff et al., 2008) | 135 | | |
|     Logit Model (Andreff et al., 2008) | 204 | | |
|     Hurdle Model (Scelles et al., 2020) | | | 139 |
|     Tobit Model (Scelles et al., 2020) | | | 138 |
|     Tobit Model (Maennig and Wellbrock, 2008) | 153 | | |
|     OLS (Celik and Gius, 2014) | | 104 | |

Applying the algorithm in the context of Tokyo 2020, we find that there will be no movement on the top of the medal count compared to the 2016 Olympics (cf. Table 4). While the United States are predicted to defend their top position, the distance to the pursuers China (+17 medals compared to 2016), Great Britain (+7), and Russia (+7) should melt.

**Table 4.**

Forecast medal count of the Olympic Games Tokyo 2020 (scenarios with and without COVID-19) and comparison to the actual results Rio De Janeiro 2016.

| Rank | Nation | Medal Forecast | Medal Forecast (No COVID-19) | Delta | Medals 2016 | Delta |
|---|---|---|---|---|---|---|
| 1 | United States | 120 | 120 | 0 | 121 | -1 |
| 2 | China | 87 | 85 | +2 | 70 | +17 |
| 3 | Great Britain | 74 | 71 | +3 | 67 | +7 |
| 4 | Russia | 63 | 62 | +1 | 56 | +7 |
| 5 | Japan | 51 | 50 | +1 | 41 | +10 |
| 6 | Germany | 45 | 44 | +1 | 42 | +3 |
| 7 | France | 44 | 42 | +2 | 42 | +2 |
| 8 | Italy | 32 | 32 | 0 | 28 | +4 |
| 9 | Australia | 29 | 30 | -1 | 29 | 0 |
| 10 | Canada | 20 | 19 | +1 | 22 | -2 |
| 11 | South Korea | 20 | 19 | +1 | 21 | -1 |
| 12 | Netherlands | 19 | 19 | 0 | 19 | 0 |
| 13 | Spain | 17 | 19 | -2 | 17 | 0 |
| 14 | Hungary | 16 | 16 | 0 | 15 | +1 |
| 15 | Kazakhstan | 16 | 16 | 0 | 18 | -2 |
| 16 | New Zealand | 16 | 16 | 0 | 18 | -2 |
| 17 | Azerbaijan | 14 | 15 | -1 | 18 | -4 |
| 18 | Uzbekistan | 14 | 14 | 0 | 13 | +1 |
| 19 | Brazil | 13 | 15 | -2 | 19 | -6 |
| 20 | Denmark | 13 | 12 | 1 | 15 | -2 |

To understand the main drivers behind the forecasts better, we use the explanatory SHAP value (Lundberg and Lee, 2017). SHAP stands for "Shapley Additive Explanations" and quantifies the importance of features for a forecast. The SHAP value of one feature describes "the change in the expected model prediction when conditioning on that feature" (Lundberg and Lee, 2017, p. 5); starting from the base value, i.e. the prediction without the knowledge of any features, the combination of all SHAP values then leads to the full model forecast (Lundberg and Lee, 2017). The game-theory-based algorithm dates back to Shapley (1953) and runs in polynomial time (Lundberg et al., 2020). The most important features in our model are the number of medals won at the previous Olympic Games, the categorical variable representing the team size (more than 149 athletes), and the normalized GDP (cf. Figure 2). All of them, generally, have a positive impact on the number of medals won.

**Figure 2.**

Feature importance of the Two-Staged Random Forest.

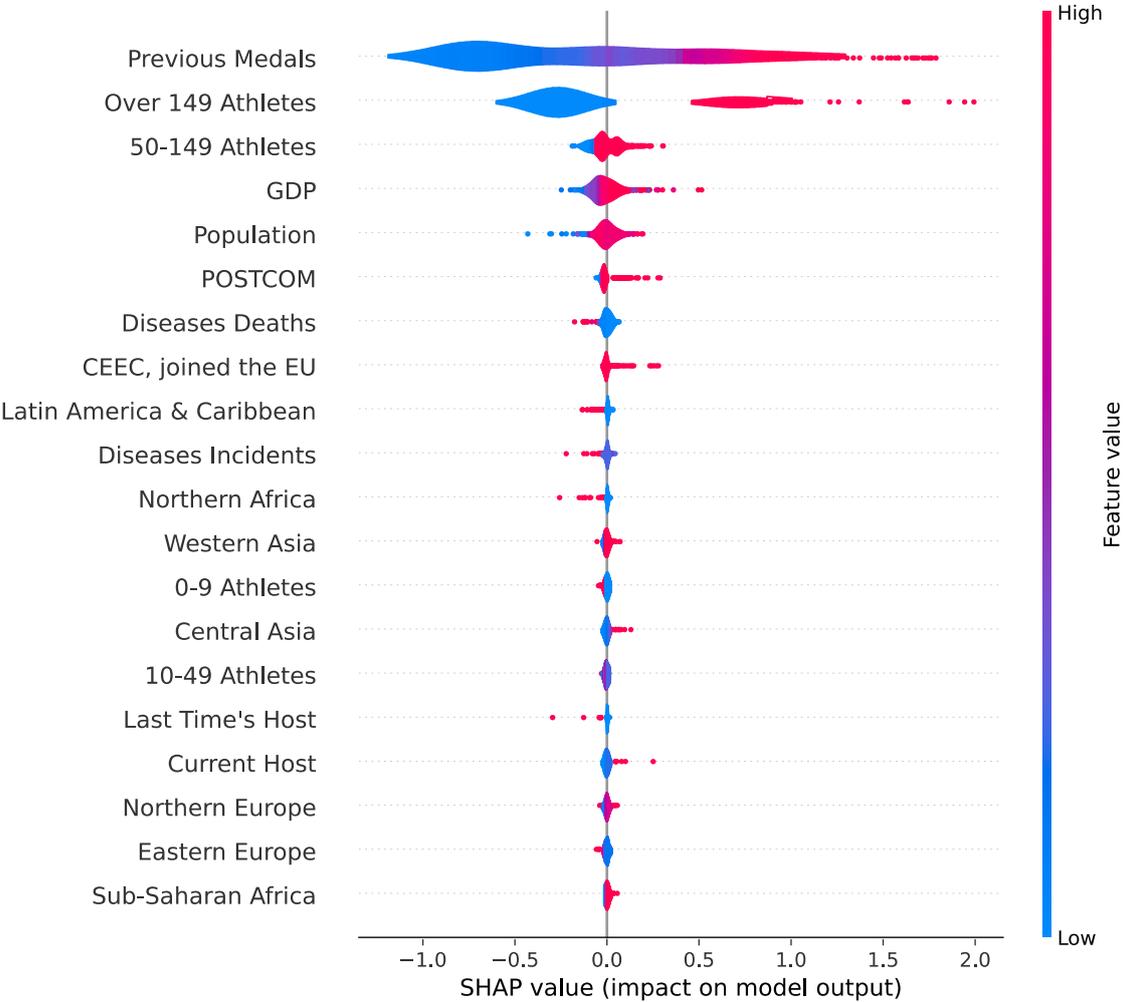

*Abbreviations and notes.* Only the 20 most relevant features are depicted. One dot represents one observation in the training data, i.e. one Olympia-nation-combination. Variables are ranked in descending order according to their feature importance. The horizontal location shows whether the effect of the value is associated with a higher or lower prediction. Color shows whether that variable is high (in red) or low (in blue) for each observation. A high "Number of Medals at previous Olympics" has a high and positive impact on the number of medals at the current Olympics. The "high" comes from the red color, and the "positive" impact is shown on the X-axis. Similarly, the "Diseases Deaths" is negatively correlated with the dependent variable.

Three of the features are directly impacted by COVID-19: GDP, incidents of and deaths from lower respiratory diseases. This allows us to create a theoretical scenario without the presence of the pandemic, such that we can clearly quantify its impact (cf. Table 4). Although, all three features significantly impact the number of medals, we find that there is little movement caused by COVID-19 amongst the top-20 nations. Notable, however, is the severe impact of the pandemic on the American economy and health system causing a further reduction of the advance of the United States although absolute medal figures remain largely constant.

We expect the highest gains in China (+2 medals). Although the pandemic originated in China, the country was hit less severe in global comparison. Liu et al. (2021) mention a high media coverage and an efficient contact tracing as success factors of the Chinese government in fighting COVID-19. We illustrate the main drivers of this development (cf. Figure 3). Compared to the no COVID-19 scenario, China could slightly increase its share in the world's GDP versus weaker economies around the globe. On a practical level, this means that training measures and competitions as preparation for Tokyo 2020, do not have to be cancelled unexpectedly to cover funds in other areas. Both incidents of and deaths from lower respiratory diseases remained on a low level. Altogether, we experience no change in the medal forecast for the Chinese team. However, as the total number of medals forecast by the model declines, scaling moves China up in the medal count.

**Figure 3.**

Individual feature importance for China in the current scenario (top) and in the no COVID-19 scenario (bottom).

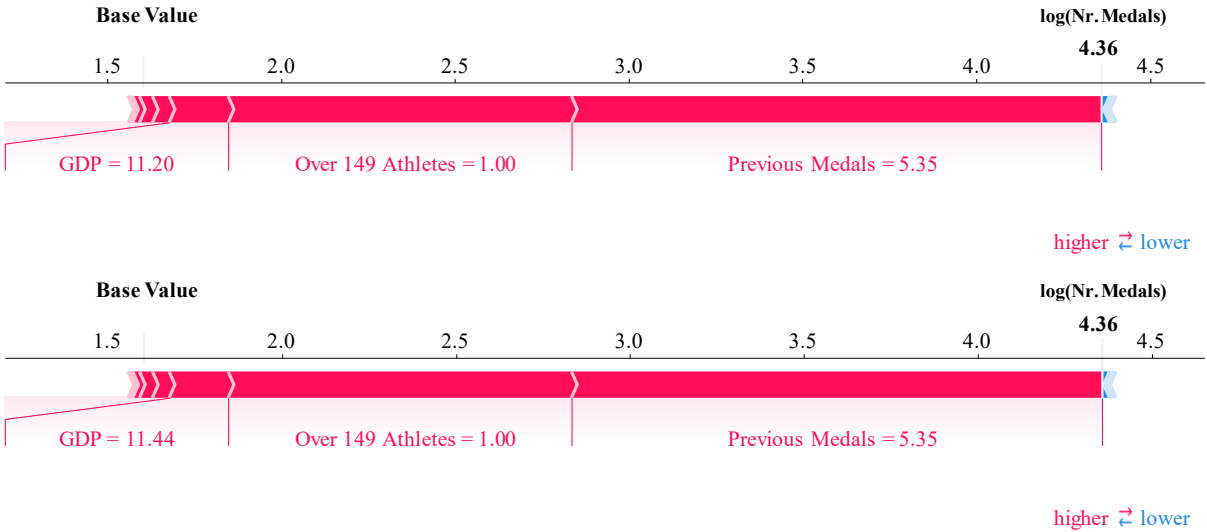

*Abbreviations and notes.* The value for log(Nr. Medals) describes the respective forecast. The base value would be predicted without any knowledge for the current output. Features that push the prediction higher, i.e. to the right are shown in red, while those pushing the prediction lower are illustrated in blue.

On the opposite, Spain experiences a loss of two medals. Both, a relatively smaller GDP and an increased number of incidents of lower respiratory diseases are responsible for this development (cf. Figure 4).

**Figure 4.**

Individual feature importance for Spain in the current scenario (top) and in the no COVID-19 scenario (bottom).

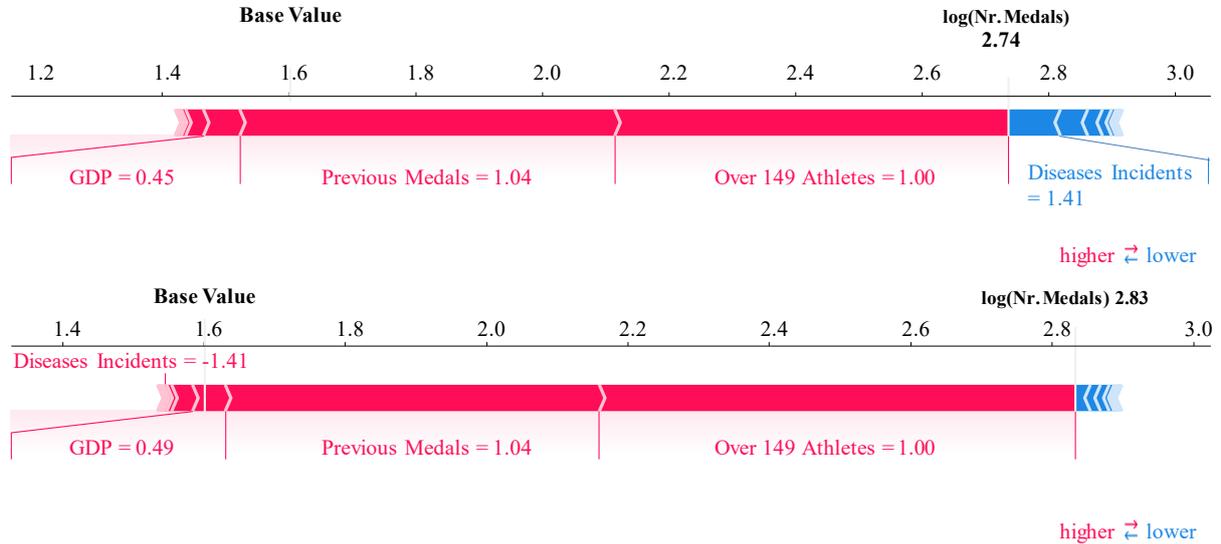

*Abbreviations and notes.* The value for log(Nr. Medals) describes the respective forecast. The base value would be predicted without any knowledge for the current output. Features that push the prediction higher, i.e. to the right are shown in red, while those pushing the prediction lower are illustrated in blue.

## 4. Discussion

Applying a two-staged Random Forest, we significantly improved the forecasting accuracy regarding Olympic medals outperforming a naïve model and currently existing statistical approaches applied in academic literature. A forecast of the Tokyo 2020 Olympic Games hosted in 2021 suggests that the United States lead the medal count followed by China and Great Britain. Particularly China, that has largely invested in sports development, is likely to exhibit a rise in medals. Our findings are highly relevant for several stakeholders, not only ex ante, i.e. before the Olympics, but also ex post. Ex ante, media companies and sport sponsors could allocate their resources to promising nations, that are likely to increase their performance compared to the previous Olympics (e.g. China). While spectators demand stories about Olympic heroes rather than group-stage knock-outs, the right focus when planning documentaries or interviews is essential for the media to reach a high audience share. The same concept holds for sponsors who profit from signing athletes, that are at the centre of high media attention. Sports betting companies offer bets on the Olympic medal count. While they generally apply different datasets than the one used in this paper to determine the odds, the strong performance of the two-staged Random Forest suggests that a detailed comparison of both models and potential re-calibration might be beneficial.

Ex post, sports politicians and managers are facing the challenge to judge the performance of their teams. Our forecast allows them to detect over- or underperformance against what was to be expected ex ante. Such an evaluation helps to assess the impact of specific investments or training concepts. Subsequently, funds for preparing the team for the next Olympics can be allocated. The forecast shows that COVID-19 hardly impacts the number of medals among the top-20 nations. This is mainly due to the fact that decision trees generally (and hence the two-staged Random Forest) exhibit weaknesses regarding extrapolation, in this case caused by the surge in incidents of and deaths from COVID-19 (e.g. Zhao et al., 2020). Training our model with data of the Tokyo 2020 Olympic Games allows us to quantify the impact of a pandemic like COVID-19 even better. Only then, policy leaders will get a reliable picture on the connection between the management of a pandemic and national sportive success. Two ways to further improve the performance of the model are the inclusion of additional features and a novel approach for missing data points:

First, socio-economic, e.g. investments in sports infrastructure, athlete-specific, e.g. age or disciplines of athletes, and COVID-specific, e.g. number of cancelled national sports events, deliver additional insights and, thus, might improve the forecasting accuracy. Brown et al. (2018) use social media data to forecast football matches, which is an approach that could be applied to Olympic Games as well. However, as machine learning methodologies are prone to overfitting, adding new features is only possible to some extent.

Second, while we use inter- and extrapolation to handle missing data points, Hassan et al. (2009) generate the missing values using their probability distribution function. This approach outperforms the conventional mean-substitution approach, however, superiority to inter- and extrapolation, as applied in this paper, still needs to be proven.

Besides working on model-specific adjustments, scholars can build upon our research within the scope of new applications in sports forecasting. As the Olympic Games are not the only important global sports event, both, the comprehensive data set and concept of the two-staged Random Forest, presented in this paper, can be leveraged in the context of other competitions, e.g. the Football World Cup, as well.

# Appendix

**Table A1**

Underlying models of published Olympic forecasts.

| Authors | Data sample | Summer / Winter | OLS | Binary | Poisson | Tobit | Two-step | Other model |
|---|---|---|---|---|---|---|---|---|
| Ball (1972) | 1964 | S | | | | | | X |
| Grimes et al. (1974) | 1936, 1972 | S | | | | X | | |
| Baimbridge (1998) | 1896-1996 | S | X | | | | | |
| Condon et al. (1999) | 1996 | S | X | | | | | X |
| Kuper and Sterken (2001) | 1896-2000 | S | X | | | | | |
| Hoffmann et al. (2002) | 2000 | S | X | | | | | |
| Tcha and Pershin (2003) | 1988-1996 | S | | | | X | | |
| Johnson and Ali (2004) | 1952-2000 | S, W | X | X | | | | |
| Bernard and Busse (2004) | 1960-1996 | S | | | | X | | |
| Lui and Suen (2008) | 1952-2004 | S | | | X | X | | |
| Andreff et al. (2008) | 1976-2004 | S | | X | | X | | |
| Maennig and Wellbrock (2008) | 1960-2004 | S | | | | X | | |
| Forrest et al. (2010) | 1996-2004 | S | | | | X | | |
| Leeds and Leeds (2012) | 1996-2008 | S | | | X | | | |
| Vagenas and Vlachokyriakou | 2004 | S | X | | | | | |
| Emrich et al. (2012) | 1996-2010 | S, W | X | | | | | |
| Celik and Gius (2014) | 1996-2008 | S | X | | | | | |
| Trivedi and Zimmer (2014) | 1988-2012 | S | | | | | X | |
| Forrest et al. (2015) | 1960-2008 | S | | | | X | | |
| Lowen et al. (2016) | 1996-2012 | S | | | | X | | |
| Noland and Stahler (2016b) | 1960-2012 | S | X | | | X | | X |
| Noland and Stahler (2016a) | 1960-2012 | S, W | | | | X | | |
| Noland and Stahler (2017) | 1960-2012 | S, W | | | | X | | |
| Blais-Morisset et al. (2017) | 1992-2012 | S | | | X | | | |
| Forrest et al. (2017) | 1992-2012 | S | | | | X | | |
| Vagenas and Palaiothodorou (2019) | 1996-2016 | S | | | | X | | |
| Scelles et al. (2020) | 1992-2016 | S | | | | X | X | |
| Rewilak (2021) | 1996-2016 | S | | | | X | X | |

*Abbreviations and notes.* Ordinary least squares (OLS); Binary Probit / Logit regression (Binary); Poisson-based model (Poisson); Tobit model (Tobit); Two-step / Hurdle model (Two-step); Summer Games (S); Winter Games (W).

**Table A2**

Complete forecast medal count of the Olympic Games Tokyo 2020 including 95% confidence intervals (scenarios with and without COVID-19).

| Rank | Nation | Medal Forecast | Min Confidence | Max Confidence | Medal Forecast (No COVID-19) | Min Confidence (No COVID-19) | Max Confidence (No COVID-19) | Delta COVID-19 vs. no COVID-19 |
|---|---|---|---|---|---|---|---|---|
| 1 | United States | 120 | 111.0 | 131.8 | 120 | 116.5 | 128.1 | 0 |
| 2 | China | 87 | 79.5 | 94.9 | 85 | 78.6 | 95.9 | +2 |
| 3 | Great Britain | 74 | 68.6 | 80.8 | 71 | 67.0 | 77.2 | +3 |
| 4 | Russia | 63 | 55.6 | 70.8 | 62 | 56.5 | 70.6 | +1 |
| 5 | Japan | 51 | 43.6 | 58.7 | 50 | 43.6 | 58.7 | +1 |
| 6 | Germany | 45 | 42.4 | 47.6 | 44 | 42.7 | 47.2 | +1 |
| 7 | France | 44 | 38.9 | 48.6 | 42 | 40.1 | 46.7 | +2 |
| 8 | Italy | 32 | 28.7 | 37.0 | 32 | 29.5 | 35.9 | 0 |
| 9 | Australia | 29 | 25.8 | 32.5 | 30 | 28.3 | 33.9 | -1 |
| 10 | Canada | 20 | 17.3 | 22.7 | 19 | 17.9 | 21.2 | +1 |
| 11 | South Korea | 20 | 18.5 | 21.1 | 19 | 18.5 | 21.1 | +1 |
| 12 | Netherlands | 19 | 15.5 | 24.3 | 19 | 17.7 | 22.3 | 0 |
| 13 | Spain | 17 | 12.7 | 23.6 | 19 | 17.4 | 20.6 | -2 |
| 14 | Hungary | 16 | 15.1 | 17.4 | 16 | 14.7 | 17.4 | 0 |
| 15 | New Zealand | 16 | 14.2 | 18.3 | 16 | 14.2 | 18.3 | 0 |
| 16 | Kazakhstan | 16 | 13.2 | 19.6 | 16 | 13.7 | 19.7 | 0 |
| 17 | Azerbaijan | 14 | 12.7 | 16.3 | 15 | 13.2 | 16.9 | -1 |
| 18 | Uzbekistan | 14 | 12.8 | 16.6 | 14 | 12.8 | 16.3 | 0 |
| 19 | Brazil | 13 | 8.4 | 20.3 | 15 | 11.6 | 20.8 | -2 |
| 20 | Kenya | 13 | 11.5 | 15.1 | 13 | 12.0 | 15.2 | 0 |
| 21 | Denmark | 13 | 10.8 | 14.9 | 12 | 10.9 | 14.6 | +1 |
| 22 | Cuba | 12 | 9.9 | 13.8 | 11 | 9.9 | 13.8 | +1 |
| 23 | Poland | 11 | 9.8 | 12.7 | 11 | 10.0 | 12.9 | 0 |
| 24 | Jamaica | 11 | 9.0 | 12.6 | 10 | 9.0 | 12.6 | +1 |
| 25 | Serbia | 10 | 8.8 | 11.8 | 10 | 8.1 | 12.0 | 0 |
| 26 | Belarus | 10 | 8.6 | 11.4 | 10 | 8.3 | 11.6 | 0 |
| 27 | Ukraine | 10 | 7.3 | 13.3 | 11 | 9.9 | 13.7 | -1 |
| 28 | Czech Republic | 9 | 7.9 | 11.0 | 10 | 8.2 | 11.6 | -1 |
| 29 | Ethiopia | 9 | 8.0 | 10.5 | 9 | 8.0 | 10.0 | 0 |
| 30 | Croatia | 9 | 7.8 | 10.8 | 9 | 7.8 | 10.8 | 0 |
| 31 | Sweden | 9 | 6.6 | 12.1 | 10 | 9.1 | 11.5 | -1 |

| Rank | Nation | Medal Forecast | Min Confidence | Max Confidence | Medal Forecast (No COVID-19) | Min Confidence (No COVID-19) | Max Confidence (No COVID-19) | Delta COVID-19 vs. no COVID-19 |
|---|---|---|---|---|---|---|---|---|
| 32 | Georgia | 8 | 6.9 | 10.3 | 8 | 7.5 | 9.8 | 0 |
| 33 | South Africa | 8 | 5.3 | 13.3 | 11 | 8.9 | 13.7 | -3 |
| 34 | Switzerland | 7 | 5.8 | 9.6 | 8 | 6.5 | 9.9 | -1 |
| 35 | Turkey | 7 | 5.3 | 8.9 | 7 | 5.5 | 8.7 | 0 |
| 36 | Colombia | 7 | 4.4 | 10.5 | 10 | 8.5 | 12.2 | -3 |
| 37 | North Korea | 6 | 5.5 | 7.3 | 6 | 5.5 | 7.3 | 0 |
| 38 | Iran | 6 | 5.1 | 7.3 | 6 | 5.2 | 7.1 | 0 |
| 39 | Thailand | 5 | 4.6 | 5.8 | 5 | 4.6 | 5.8 | 0 |
| 40 | Greece | 5 | 4.3 | 6.4 | 5 | 4.1 | 6.3 | 0 |
| 41 | Belgium | 5 | 3.5 | 6.6 | 6 | 4.7 | 6.6 | -1 |
| 42 | Chinese Taipei | 5 | 3.9 | 5.3 | 4 | 3.9 | 5.3 | +1 |
| 43 | Slovakia | 4 | 3.8 | 5.2 | 4 | 3.5 | 5.0 | 0 |
| 44 | Malaysia | 4 | 3.3 | 5.8 | 4 | 3.3 | 5.8 | 0 |
| 45 | Lithuania | 4 | 3.7 | 5.2 | 4 | 3.7 | 5.1 | 0 |
| 46 | Venezuela | 4 | 3.2 | 5.8 | 4 | 3.2 | 5.8 | 0 |
| 47 | Armenia | 4 | 3.5 | 5.2 | 5 | 4.2 | 5.6 | -1 |
| 48 | Romania | 4 | 3.5 | 5.3 | 4 | 3.8 | 5.5 | 0 |
| 49 | Slovenia | 4 | 3.6 | 4.9 | 4 | 3.7 | 4.8 | 0 |
| 50 | Norway | 4 | 3.7 | 4.8 | 4 | 3.7 | 4.8 | 0 |
| 51 | Bulgaria | 4 | 3.5 | 4.8 | 4 | 3.1 | 4.5 | 0 |
| 52 | Indonesia | 4 | 3.2 | 5.1 | 4 | 3.5 | 5.1 | 0 |
| 53 | Mexico | 4 | 2.9 | 5.1 | 5 | 4.3 | 6.0 | -1 |
| 54 | Tunisia | 4 | 3.1 | 4.3 | 4 | 3.1 | 4.3 | 0 |
| 55 | India | 4 | 2.1 | 5.8 | 4 | 2.1 | 6.1 | 0 |
| 56 | Argentina | 4 | 2.4 | 5.2 | 6 | 4.6 | 7.7 | -2 |
| 57 | Algeria | 3 | 2.5 | 4.1 | 3 | 2.1 | 3.7 | 0 |
| 58 | Vietnam | 3 | 2.3 | 4.1 | 2 | 1.7 | 3.0 | +1 |
| 59 | Egypt | 3 | 1.9 | 4.2 | 3 | 2.0 | 4.4 | 0 |
| 60 | Ireland | 3 | 2.3 | 3.6 | 3 | 2.4 | 3.3 | 0 |
| 61 | Mongolia | 3 | 2.1 | 3.5 | 3 | 2.1 | 3.5 | 0 |
| 62 | Philippines | 3 | 2.0 | 3.4 | 3 | 1.9 | 3.4 | 0 |
| 63 | Nigeria | 2 | 1.9 | 3.2 | 2 | 1.8 | 3.2 | 0 |
| 64 | Latvia | 2 | 2.1 | 2.7 | 2 | 2.1 | 2.7 | 0 |
| 65 | Israel | 2 | 1.8 | 3.1 | 2 | 1.7 | 2.8 | 0 |
| 66 | Finland | 2 | 1.9 | 2.9 | 2 | 1.9 | 3.0 | 0 |
| 67 | Estonia | 2 | 1.9 | 2.8 | 2 | 1.9 | 2.5 | 0 |
| 68 | Morocco | 2 | 2 | 3 | 2 | 1.6 | 2.1 | 0 |

| Rank | Nation | Medal Forecast | Min Confidence | Max Confidence | Medal Forecast (No COVID-19) | Min Confidence (No COVID-19) | Max Confidence (No COVID-19) | Delta COVID-19 vs. no COVID-19 |
|---|---|---|---|---|---|---|---|---|
| 69 | Trinidad and Tobago | 2 | 1.8 | 2.4 | 2 | 1.8 | 2.4 | 0 |
| 70 | Bahrain | 2 | 1.7 | 2.3 | 2 | 1.6 | 2.2 | 0 |
| 71 | Portugal | 2 | 1.6 | 2.2 | 2 | 1.7 | 2.3 | 0 |
| 72 | Austria | 2 | 1.6 | 2.1 | 2 | 1.6 | 2.2 | 0 |
| 73 | Ivory Coast | 2 | 1.5 | 2.1 | 2 | 1.4 | 2.0 | 0 |
| 74 | Fiji | 2 | 1.5 | 2.1 | 2 | 1.5 | 2.1 | 0 |
| 75 | Kyrgyzstan | 2 | 1.5 | 1.8 | 0 | | | +2 |
| 76 | Tajikistan | 2 | 1 | 2 | 2 | 1 | 2 | 0 |
| 77 | Singapore | 2 | 1.4 | 1.8 | 2 | 1 | 2 | 0 |
| 78 | Bahamas | 2 | 1.3 | 1.8 | 2 | 1 | 2 | 0 |
| 79 | Moldova | 2 | 1.3 | 1.8 | 2 | 1 | 2 | 0 |
| 80 | Kosovo | 1 | 1 | 2 | 2 | 1 | 2 | -1 |
| 81 | Grenada | 1 | 1 | 1 | 1 | 1 | 1 | 0 |
| 82 | Guatemala | 0 | | | 1 | 1 | 1 | -1 |

*Abbreviations and notes.* Confidence intervals are computed by grouping the 1,000 decision trees of the Random Forest in 100 groups of 10 decision trees. Within these groups serving as smaller Random Forests mean values are computed. The 100 obtained data points are reduced by eliminating the five data points with the greatest deviation from the mean. The remaining 95 values determine the 95% confidence interval. Confidence intervals are only computed for nations with predicted medal success. All nations from rank 83 on have zero values.